\documentclass[12pt]{article}

\usepackage{graphicx}
 \usepackage{color}
\usepackage{epsfig}

\usepackage{graphics}

\usepackage{anysize}
\usepackage{amsmath,amsfonts, amssymb,amsthm}

\begin{document}

\title{A nonclassical symbolic theory of working memory, mental
computations, and mental set}

\author{Victor Eliashberg\thanks{ The research is supported by the
Grant FA9550-08-1-0129 under the DARPA DSO Mathematical Challenge
One: The Mathematics of
 the Brain. Proposal \cite{bib34}.} \\ Department of Electrical Engineering
\\ Stanford  University}

\date{} \maketitle

\begin{abstract} The paper tackles four basic questions associated
with human brain as a learning system. How can the brain learn to
(1) mentally simulate different external memory aids, (2) perform,
in principle, any mental computations using imaginary memory aids,
(3) recall the real sensory and motor events and synthesize a
combinatorial number of imaginary events, (4) dynamically change its
mental set to match a combinatorial number of contexts? We propose a
uniform answer to (1)-(4) based on the general postulate that the
human neocortex processes symbolic information in a ``nonclassical"
way. Instead of manipulating symbols in a read/write memory, as the
classical symbolic systems do, it manipulates the states of
dynamical memory  representing different temporary attributes of
immovable symbolic structures stored in a long-term memory. The
approach is formalized as the concept of E-machine. Intuitively, an
E-machine is a system that deals mainly with characteristic
functions representing subsets of memory pointers rather than the
pointers themselves. This nonclassical symbolic paradigm is Turing
universal, and, unlike the classical one, is efficiently
implementable in homogeneous neural networks with temporal
modulation topologically resembling that of the neocortex.

\end{abstract}
 \section{Introduction} \label{sec1}

Conventional computers process symbolic information by manipulating
symbols in a read/write memory -- call it a RAM buffer. This
classical symbolic computational paradigm encounters serious
problems as a metaphor for the symbolic level of information
processing in the human brain. It is unlikely that the brain has a
counterpart of a conventional RAM buffer. Even if such a buffer
existed it would be too small and too slow to allow the brain to
efficiently process symbolic information in a traditional way.

\textbf{Q0.} \emph{How can the brain produce such cognitive
phenomena as working memory, mental computations, and language
without a RAM buffer?} To tackle this general question, it is
helpful to start with the following observations:

 \medskip \noindent \textbf{1. Working memory}. People learn
to mentally simulate different external memory aids with the
properties of a read/write memory. For example, a highly skilled
abacus user learns to compute on an imaginary abacus as efficiently
as on the real device \cite{bib3}. Similarly, an experienced chess
player learns to play chess on an imaginary chess board. A computer
simulation of a chess board requires a RAM buffer with no less than
$n=64$ addresses and $m=13$ symbols -- twelve symbols representing
chess pieces of two colors and one symbol representing an empty
square. This observation raises the question:

\textbf{Q1}. \emph{How can a  system learn to simulate a RAM
buffer?}

\noindent We show that no learning system that uses a
gradient-descent-type or a statistical-optimization-type learning
algorithm can learn to simulate even a small RAM buffer with $n=m=2$
(Theorems 1 and 2 of Section \ref{sec4}). Though a RAM buffer is a
deterministic system, its behavior is statistically unpredictable in
the traditional sense. This means that the human brain  cannot rely
on the traditional statistical prediction techniques to mentally
simulate the behavior of the external world.

\medskip \noindent \textbf{2. Turing universality and learning}. A
person with a good visual memory can be taught to perform, in
principle, any mental computation with the use of an imaginary
memory aid. Ignoring some theoretically unimportant limitations on
the size of the imaginary memory aid, this observation means that
the human brain must be treated by a system theorist as a Turing
universal learning system. It is interesting to ask:

\textbf{Q2.} \emph{What is the simplest architecture of a Turing
universal learning system?}

\noindent This question is directly related to Q1. It is easy to
prove that a learning system that cannot answer question Q1 cannot
be a Turing universal learning system.

\medskip \noindent \textbf{3. Memorization, recollection, and
synthesis}. People can memorize and recall long sequences of real
sensory and motor events. At the same time, they can synthesize a
combinatorial number of imaginary events. It is attractive to think
that the same learning algorithm can account for all outlined
phenomena. We can ask:

 \textbf{Q3}. \emph{What learning algorithm satisfies the requirements of
 correct recollection, and combinatorial synthesis?}

\noindent We argue that a learning algorithm that attempts to do a
lot of preprocessing of the learner's  experience before putting
this experience in the learner's LTM cannot answer this question. In
contrast, an algorithm that simply memorizes all learner's ``raw''
experience, call it a \emph{complete memory algorithm} (CMA), does
not have this limitation (Section 6).

  \medskip \noindent \textbf{4. Mental set and context.} People interpret their
inputs differently depending on context. The number of possible
contexts explodes exponentially. This leads to a question:

\textbf{Q4.} \emph{How can a system with a linearly growing size of
knowledge learn to efficiently deal with an exponentially growing
number of possible contexts?}

\noindent This problem of ``Attitude and Context'' \cite{bib33}
calls into question all traditional approaches to brain modeling and
cognitive modeling . The brain cannot have different mental agents
\cite{bib18b} for all possible contexts. Accordingly, a consistent
answer to Q4 must explain how the brain can dynamically synthesize
such agents depending on context.

\medskip  This paper offers a uniform  answer  to questions Q1-Q4
based on the assumption that the human neocortex processes symbolic
information in a ``nonclassical" way. We postulate that, instead of
manipulating symbols in a read/write memory, as the classical
symbolic systems do, the brain manipulates the states of dynamical
memory representing different temporary attributes of immovable
symbolic structures stored in a long-term memory. This integrated
symbolic-dynamical  paradigm is referred to as the concept of
E-machine \cite{bib7,bib8,bib9}. Answering questions Q1-Q4 provides
an insight into the general question Q0.

A complex E-machine (CEM) is a hierarchical associative learning
system built from several (many) primitive associative learning
systems called primitive E-machines (PEM) (Section \ref{sec5}). A
PEM has two main types of states:

\noindent (1) The states of encoded long-term memory (LTM)
representing symbolic knowledge (software). These states are called
\emph{G-states}, the `G' implying the notion of ``synaptic Gain''.
(2) The states of different types of dynamical short-term memory
(STM) and intermediate-term memory (ITM) accounting for a
context-dependent dynamic reconfiguration of the knowledge stored in
LTM. These states are called  \emph{E-states}, the `E' implying the
notion of ``residual Excitation''.  The name ``E-machine''
emphasizes the special importance of the E-states.

We demonstrate some possibilities of the E-machine paradigm by
presenting an example of a learning robot with the brain organized
as an E-machine. The robot interacts with an external world
represented by a keyboard and a screen. The model (implemented as an
interactive C++ program) produces the following educational effects:

(1) \emph{Working memory}. After interacting with an external
read/write memory aid the robot learns to mentally simulate this
memory aid (creates an imaginary memory aid).

(2) \emph{Mental computations}. The robot learns to perform
computations with the use of the imaginary memory aid.

(3) \emph{Mental set and context}. The robot interprets its visual
input in a combinatorial number of different ways depending on the
context created through an auditory input. The rest of the paper
consists of the following sections:
\\

2. System (Robot,World) as a cognitive model.

3. The external world as a generalized RAM (GRAM).

4. Limitations of traditional learning systems.

5. The concept of a primitive E-machine (PEM).

6. A PEM can learn to simulate a GRAM.

7. On Turing universality and learning.

8. The effect of a context-dependent mental set.

9. Promising directions of research.

10. Methodological remarks.

 \section{System (Robot,World) as a cognitive model} \label{sec2}

\begin{figure}[ht!] \begin{center}
\includegraphics[width=3.4in]{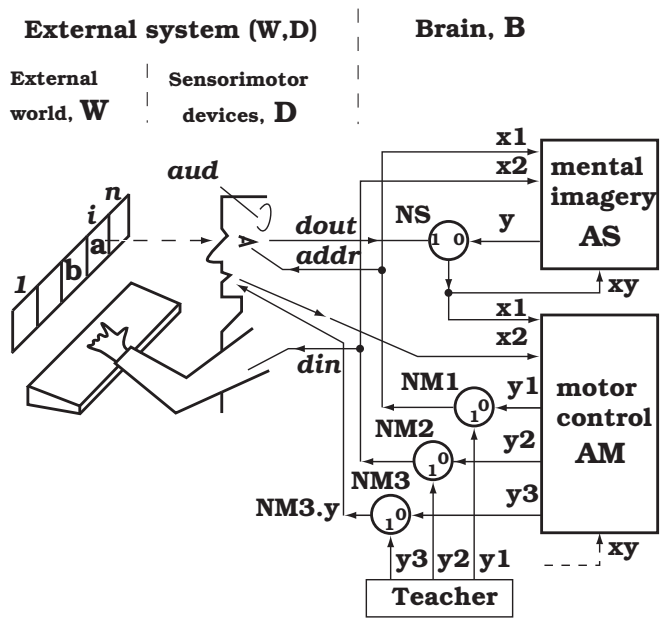}
 \end{center}
\caption{Experimental system } \label{fig1} \end{figure}

The general architecture of the cognitive model used in this paper
is shown in Figure \ref{fig1}. The model consists of an external
world, W (represented by a keyboard and a screen), and a robot,
(D,B), consisting of the sensorimotor devices, D, and the brain, B.
 From the system-theoretical viewpoint, it is convenient to treat
system (W,D,B) as a composition of two subsystems: the
\emph{external system}, (W,D) and the brain B. In this
representation, both systems can be viewed as abstract machines, the
outputs of (W,D) being the inputs of B, and vice versa. Note that
the brain does not know about the external world, W, per se. It
knows only about the external system (W,D).

The screen is divided into squares. For simplicity, only one row of
squares is shown. For now, we  assume that the robot's eye can see
only one square at a time, call it the \emph{scanned square} -- the
idea is borrowed from Turing \cite{bib30}. We also assume that the
system has some eye tracking device (not shown), so when the robot
depresses a key the character appears in the scanned square.

The external system, (W,D), behaves, essentially, as a RAM. Two
\emph{motor} inputs, $addr$ and $din$, representing the eye position
and the typed character, serve, respectively, as the \emph{address}
and \emph{data} input of the RAM. The \emph{ sensory} (visual)
output, $dout$, serves as the \emph{data} output. There is no
\emph{control input} similar to \emph{write-enable} input of a RAM.
We can say that system (W,D) is in \emph{write mode} when there is a
\emph{nonempty} motor input, $din$. Otherwise, the system is in
\emph{read mode}. In Section \ref{sec3.1} we will formalize this
verbal description as the concept of a \emph{generalized RAM}
(GRAM).

The diagram  depicts a delayed speech feedback, NM3.\emph{y}
$\rightarrow$  AM.\emph{x2}, and an auditory input, $aud$ (we use ``
. '' as the   membership operator, that is, AM.\emph{x2} denotes
input \emph{x2} of unit AM). These will be needed in Sections
\ref{sec7} and \ref{sec8}. The robot's brain is divided into four
units:

\begin{description}

 \item 1. Associative learning system, AS, responsible for sensory
  memory and mental imagery.  The goal of this system is to
 learn to simulate the external system, (W,D). In this case, (W,D)
 works as  a RAM.

 \item 2. Associative learning system, AM, responsible for motor
 control. The goal of this system is to learn to simulate the teacher.

 \item 3. Sensory centers (nuclei), NS, that work as a multiplexer
 switching between the output from the eye, $dout$,
 and the output from AS, $AS.y$.

\item 4. Motor centers (nuclei), NM=(NM1,NM2,NM3), that work as a
multiplexer switching between the output of the teacher,
 $T.y=(T.y1,T.y2,T.y3)$, and the output of system AM, $AM.y=(AM.y1,AM.y2,AM.y3)$.
 We assume that each multiplexer has a select input, $sel$ (not
 shown), that can be set by the experimenter.

\end{description}

Both systems, AM and AS, are in the, so-called, supervised learning
mode. In the course of training, the teacher can produce any desired
output of centers NM. The teacher can also switch the output of
centers NS ($NS.y$) between the output of system AS ($ AS.y$) and
the output of the eye, $dout$. When $NS.y=dout$, system (W,D) serves
as the teacher for system AS. Both systems, AS and AM, have inputs
denoted as $xy$. These inputs deliver the output signals needed for
learning. Such inputs are often referred to as \emph{desired
outputs}.

\section{The external world as a generalized RAM}  \label{sec3}

\subsection{The concept of a generalized RAM (GRAM)} \label{sec3.1}

\begin{figure}[h!] \begin{center}
  \includegraphics[width=3.0in]{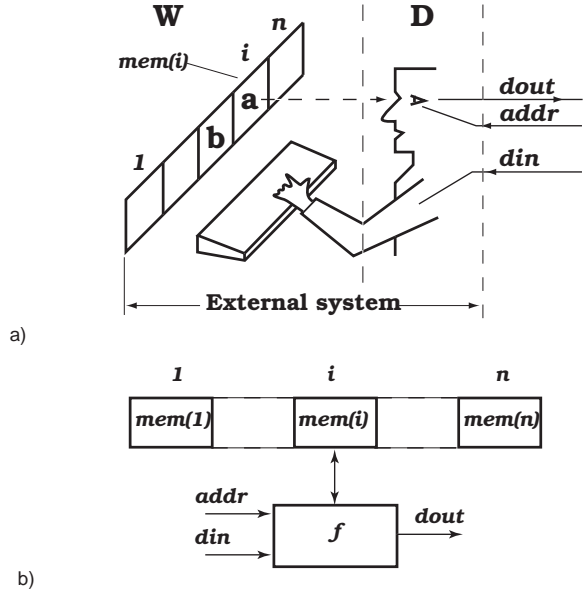}
 \end{center}
 \caption{ a) The external world appears to the brain as a ``RAM buffer''. b) The concept
 of a generalized RAM (GRAM).} \label{fig2}
\end{figure}

\textbf{Definition. }
 A \emph{generalized RAM (GRAM)} is a system
$ (\textbf{A},\textbf{D},\textbf{M},f) $ (see Figure \ref{fig2}),
where
 \begin{itemize}
 \item $\textbf{A}=\{a_1,...a_n \}$ is a set of symbols
 called \emph{address set}.
 \item $\textbf{D}=\{d_1,...d_m,\varepsilon \}$ is a set of symbols
 called \emph{data set}, where $ \varepsilon $ is the empty
 symbol meaning ``no data''.
 \item $\textbf{M}=\textbf{D} \times ...  \times \textbf{D}= \textbf{D}^n
 $ is the \emph{set of memory states} represented as memory arrays $(mem(1),...mem(n))$,
 where $mem(i) \in
 \textbf{D}$ is the data stored in the $i-th$ location of GRAM.
 \item $f:\textbf{A} \times \textbf{D} \times \textbf{M} \rightarrow \textbf{M} \times \textbf{D} $
  is a function computing the next memory state and the output data.
 \end{itemize}
 \ 

\noindent\textbf{Notation}. In what follows, we use a  MATLAB-like
notation. Array indices start with 1. Symbol `` : '' denotes the
range operator. Symbol ``$end$'' is used as a delimiter  in control
statements. We add the rightmost parentheses to include a time
index. For example, $mem(:)(\nu)$ denotes the value of $m(:)$ at the
moment $\nu$. Similarly, $dout(\nu)$ denotes the value of $dout$ at
the moment $\nu$.

\medskip \noindent In discrete time, $\nu$, the work of a GRAM is
described by expressions (1) and (2):

(1) \hspace{0.1in} $(mem(:)(\nu+1),dout(\nu)) \  = \
f(addr(\nu),din(\nu),m(:)(\nu))$ ;

\noindent where function $f$ is

(2) \hspace{0.1in} $if(din == \varepsilon) \ dout=mem(addr);  \ else
\ \{  dout=din;
 \ mem(addr)=din; \} $ $end$;
\\\\
The main difference between a GRAM and a conventional RAM is as
follows: \begin{enumerate}
 \item Both address and data are treated as symbols: that is, only
 the \emph{equal/not equal} relationship is defined for the elements of \textbf{A} and \textbf{D}.
 \item  GRAM is always in write mode when input data is present, $din \neq \varepsilon
 $. Only if input data is not present, $din=\varepsilon $, is GRAM in read mode.
 With this approach, we do not need a special control input, e.g.,
 \emph{write\_enable}, to indicate the write or read mode.
\end{enumerate}

\noindent One can further extend the notion of GRAM and allow
different sets $\textbf{D}^{in}$ and $\textbf{D}^{out}$ of  input
and output data. What we need is a (not necessarily injective) map
$\textbf{D}^{in}\to  \textbf{D}^{out}$.

 \subsection{An experiment with a GRAM} \label{sec3.2}

\begin{figure}[h!] \begin{center}
  \includegraphics[width=2.0in]{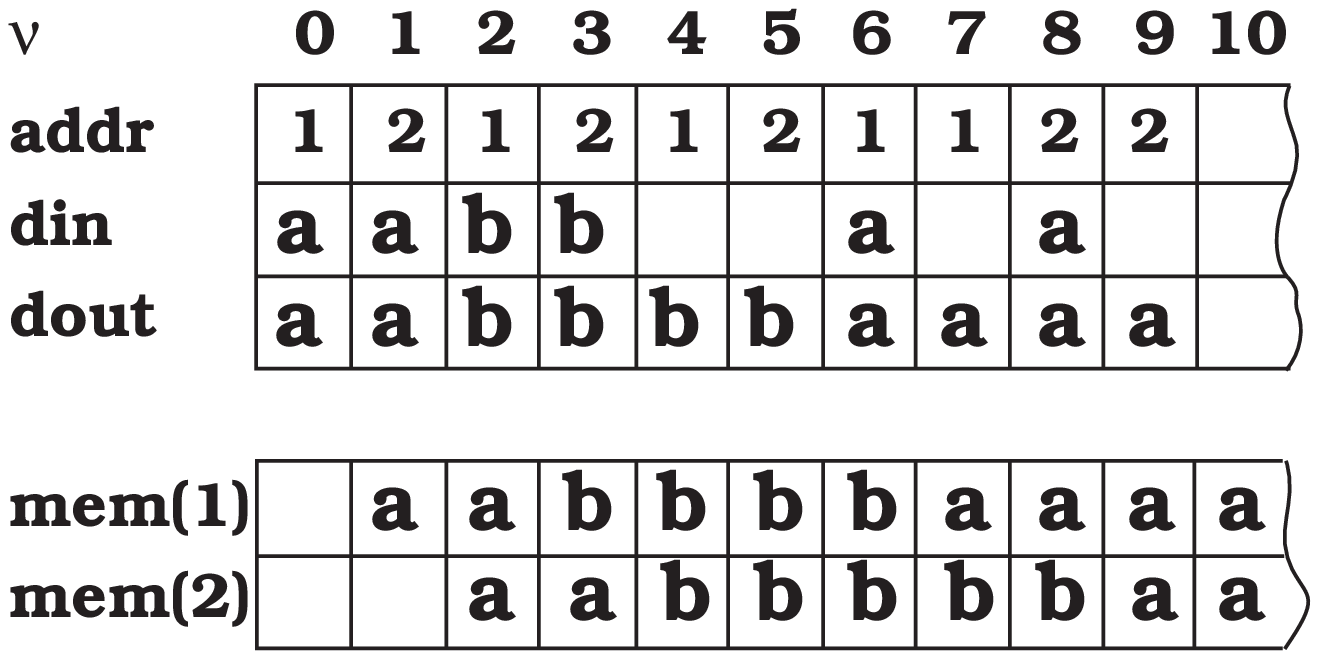}
 \end{center}
 \caption{Experiment with a GRAM. $\textbf{A}=\{1,2 \}$, $\textbf{D}=\{a,b,\varepsilon \}$.
 The empty symbol $\varepsilon $ is shown as a blank square. } \label{fig3}
\end{figure}

 To get used to the notion of GRAM, it is helpful to follow the
experiment with a GRAM shown in Figure \ref{fig3}, where
$\textbf{A}=\{1,2\}$, and $\textbf{D}=\{a,b,\varepsilon\}$. The
three-row table in the upper part of the figure displays
 the input/output sequence of GRAM as a function of discrete time
 $\nu =0,1,..9$.
  The  $din= \varepsilon$  entries are shown  as  blank
squares.  The two-row
 table in the lower part of the figure displays
 the contents of the two memory locations, $mem(1),mem(2)$ as functions of $\nu$.
 The  $mem(i)= \varepsilon$  entries are shown  as  blank squares.
 For instance, we have
\\
$\nu=0$: \\
 Memory is empty:
$mem(:)=(mem(1),mem(2))=(\varepsilon,\varepsilon)$. Input
$(addr,din)=(1,a)$ produces  output $dout=a$, and writes $a$ into
location 1, $mem(1)=a$.
 \\
 $\nu=3$:
 \\
 Memory state is:
$ mem(:)=(b,a)$. Input  $(addr,din)=(2,b)$ produces output $dout=b$,
and writes $b$ into location 2, $mem(2)=b$.
 \\
$\nu=9$:
 \\
 Memory state is:
$ mem(:)=(a,a)$. Input $(addr,din)=(2,\varepsilon)$ reads data from
location 2, $dout=mem(2)=a$. Memory state does not change.

\subsection{Fixed rules and variable rules} \label{sec3.3}

 Analyzing the three-row input/output
table shown in the upper part of Figure \ref{fig3}, we can discover
two types of $x \rightarrow y$ rules, where $x=(addr,din)$ and
$y=dout$:

\begin{enumerate}

\item Rules of the type $(addr,din \neq \varepsilon) \rightarrow
dout$, are called \emph{fixed rules}. In this specific example, the
fixed rules are: $(1,a) \rightarrow a$, $(2,a) \rightarrow a$,
$(1,b) \rightarrow b$, and $(2,b) \rightarrow b$. There are $m \cdot
n=2 \cdot2=4$ such rules. Fixed rules can be easily extracted from
the shown input/output sequence by different learning algorithms.

\item Rules of the type $(addr,din = \varepsilon) \rightarrow dout$,
are called  \emph{variable rules}. In this example, the variable
rules are: $(1,\varepsilon) \rightarrow dout$, and $(2,\varepsilon)
\rightarrow dout$. There are $n$ variable rules. The output part,
$dout$, of a variable rule depends on the most recently executed
fixed rule with the same address. For example, the output  of rule
$(1, \varepsilon) \rightarrow dout$ at $\nu=7$ is $dout=a$, because
the most recently executed fixed rule with $addr=1$ is rule $(1,a)
\rightarrow a$ at $\nu=6$. Variable rules cannot be correctly
executed by a learning system that does not save information about
the most recently executed fixed rules.
 \end{enumerate}

\noindent It is useful to view fixed rules as a tool for assigning
the right parts of variable rules. With this approach, we can say
that the \emph{meaning} or \emph{value} of an address symbol in a
variable rule depends on the most recent assignment. In the
discussed example, each address symbol from $\textbf{A}=\{1,2\}$ can
be assigned either of two meanings (data values) from $\textbf{D} -
\{\varepsilon \} = \{a,b\}$.

\section{Limitations of traditional learning systems} \label{sec4}

\begin{figure}[h!] \begin{center}
  \includegraphics[width=2.5in]{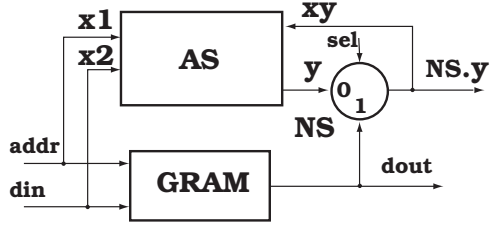}
 \end{center}
 \caption{Experimental setup for learning to simulate a GRAM } \label{fig4}
\end{figure}

Assume that a traditional learning system is used as system AS in
Figure \ref{fig1}. It is convenient to redraw the relevant part of
Figure \ref{fig1} as the experimental setup shown in Figure
\ref{fig4}, where the external system, (W,D), is replaced by GRAM.
Think of input $xy$ as the \emph{desired output} of the above
learning system.

Let $NS.sel=1$, so the GRAM serves as the target system (the
teacher) for system AS. We claim that, in this experiment of
supervised learning, no traditional learning systems, used as system
AS, can learn to simulate the target system with the properties of a
GRAM.
 In what follows we  prove this claim for two broad
classes of learning systems.
 \\\\
\textbf{Theorem 1}. Let M be a learning system with some statistical
(or any other) learning algorithm that learns to predict the output
of a target system, T, from the samples of its input/output
sequence. Let the maximum length of the samples taken into account
not exceed $m$. System M cannot learn to simulate system T with the
properties of a GRAM.

\emph{Remark}. Many learning systems treat a training sequence as a
set of input/output pairs. For such systems $m=1$.
\\\\
\emph{Proof}. We are going to show that  the reaction of a GRAM to
the input $(addr,din=\varepsilon)$ is statistically unpredictable.
It is sufficient to use a specific example of a GRAM with
$\textbf{A}=\{1,2\}$, and $\textbf{D}=\{a,b,\varepsilon\}$ discussed
in Section \ref{sec3.2}.
 A bigger GRAM can always simulate a smaller GRAM, so
the result of the theorem will hold for any GRAM with $n \geq 2$.
 Let us assume that M has learned to predict the behavior of the GRAM.
 To produce the contradiction, do the following test:

\textbf{Step 1}. Send to the input of M a sequence
$x(1),x(2)....x(m+1)$, where $x(1)=(1,a)$, $x(m+1)=(1,\varepsilon)$,
and $x(2)=...x(m)=(2,d)$, where $d \in \textbf{D}$. The reaction of
the GRAM to this sequence is $dout=a$. Suppose the reaction of M is
also $y(m+1)=a$.

\textbf{Step 2}. Send the input sequence which is the same as
before, except $x(1)=(1,b)$. By definition, system M predicts its
output from the input sequence no longer than $m$, that is, it does
not take into account $x(1)$.  The reaction of M will again be
$y(m+1)=a$. However, the reaction of the GRAM will be $dout=b$. This
proves the theorem.
 \\\\
\textbf{Definition.} Let M be a learning system with an input
alphabet \textbf{X}. Let $x(1),...x(m1),\dots,$ $ x(m2),\dots,
x(m+1)$ be an input sequence of the system M, where $1 \leq m1 < m2
\leq m$.  We will say that  M \emph{loses information about the
order of input events} if there exist $m1$ and $m2$ satisfying the
above condition such that for all
 $z1,z2 \in  \textbf{X}$, and $z1 \neq z2$, the reaction of M is the same for
 $(x(m1),x(m2))=(z1,z2)$, and  $(x(m1),x(m2))=(z2,z1)$.
\\\\
\textbf{Theorem 2}. No system M  that loses information about the
order of the input events of the target system can learn to simulate
a GRAM.
\\\\
\emph{Proof}. Let us use the  same GRAM as in Theorem 1.  Suppose M
satisfies the above definition and nevertheless has learnt to
simulate the specified GRAM. To produce the contradiction do the
following test:

\textbf{Step 1}. Send to the input of M a
sequence $x(1),...x(m1),...x(m2),...x(m+1)$, such that
$x(m1)=(1,a)$, $x(m2)=(1,b)$, $x(m+1)=(1,\varepsilon)$, and
$x(m2+1)=..x(m)=(2,d)$, where  $d \in \textbf{D}$. The reaction of
the GRAM to this sequence is $dout=b$. Suppose the reaction of M is
$y(m+1)=b$.

\textbf{Step 2}. Send the input sequence
$x(1),...x(m1),...x(m2),...x(m+1)$, that is, the same as in step 1,
but $x(m1)=(1,b)$,  and $x(m2)=(1,a)$.  The reaction of the GRAM to
this sequence is $dout=a$. However, by definition, the reaction of M
to this sequence is the same as before, that is $y(m+1)=b$. This
proves the theorem.
 \\\\
Theorems 1 and 2 show that the loss of information at the time of
learning leads to principal limitations at the time of decision
making. In the case of Theorem 1, system M attempted to predict the
output of a GRAM using the samples of the GRAM's behavior of limited
length $\leq m$. In the case of the Theorem 2, M ignored the order
of the GRAM's input events.

 \section{The concept of a primitive E-machine (PEM)}  \label{sec5}

\subsection{An abstract description}  \label{sec5.1}

A primitive E-machine (PEM) is a system
$PEM=(\textbf{X},\textbf{Y},\textbf{E},\textbf{G},f_y,f_e,f_g)$,
where \begin{itemize}
 \item \textbf{X} and \textbf{Y} are finite sets of symbols called the {\it
 input} and the {\it output} set, respectively;
 \item \textbf{G} is the set of states called the {\it states of encoded (symbolic)
 long-term memory} (LTM) or the {\it G-states}.
 As mentioned before, the letter `G' implies the notion of ``synaptic Gain''.
 The G-states  represent the symbolic knowledge (software) of an E-machine.
 \item \textbf{E} is the set of states, called the {\it states of dynamical
  short-term memory} (STM) and {\it intermediate-term memory} (ITM),  or the
    {\it E-states}.  The letter `E' implies the notion of ``residual Excitation''.
    An E-machine may have several types of E-states
    representing different temporary attributes (dynamical labels) of the
    data stored in LTM. The E-states serve as the mechanism for context-dependent
    dynamic reconfiguration of  the knowledge represented by the G-states.
 \item $f_y: \textbf{X} \times \textbf{E} \times \textbf{G} \rightarrow \textbf{Y}$
 is a function called, interchangeably, the \emph{output procedure},
 the  \emph{interpretation procedure}, or the \emph{decision making
 procedure}.\footnote{In general, $f_y$ is a probabilistic
 procedure, so the symbol ``$\rightarrow $'' should be understood
 as \emph{``compute'}', rather than  as \emph{``map''}.}
 \item $f_e: \textbf{X} \times \textbf{E} \times \textbf{G} \rightarrow \textbf{E}$
 is a function called, interchangeably, the \emph{next E-state procedure},
 or the \emph{dynamic reconfiguration procedure}.
 \item $f_e: \textbf{X} \times \textbf{E} \times \textbf{G} \rightarrow \textbf{G}$
 is a function called, interchangeably, the \emph{next G-state procedure},
 or the \emph{incremental learning algorithm}.
\end{itemize}

\noindent \emph{Remark}. A PEM may have other types of states. For
simplicity, these states are not included in the above general
description.

\medskip The work of a PEM is described in discrete time, $\nu$, as
follows:
 \\\\
(1) \hspace{0.5in} $y(\nu) = f_y(x(\nu),e(\nu),g(\nu));$
\\\\
(2) \hspace{0.5in} $e(\nu+1) = f_e(x(\nu),e(\nu),g(\nu));$
\\\\
(3) \hspace{0.5in} $g(\nu+1) = f_g(x(\nu),e(\nu),g(\nu));$
\\\\
with initial states $e(0)$ and $g(0)$,
\\\\
where $x(\nu) \in \textbf{X}$, $y(\nu) \in \textbf{Y}$, $e(\nu) \in
\textbf{E}$, and $g(\nu) \in \textbf{G}$ .
\\\\
In the next section, we present an  explicit example of a PEM.  The
model is simple enough to be theoretically understandable, but  at the
same time it is sufficiently complex to produce some nontrivial
cognitive phenomena. This PEM will be used as system AS and system
AM of Figure~\ref{fig1}.

\subsection{An example of a primitive E-machine} \label{sec5.2}

\begin{figure}[ht!] \begin{center}
 \includegraphics[width=3.5in]{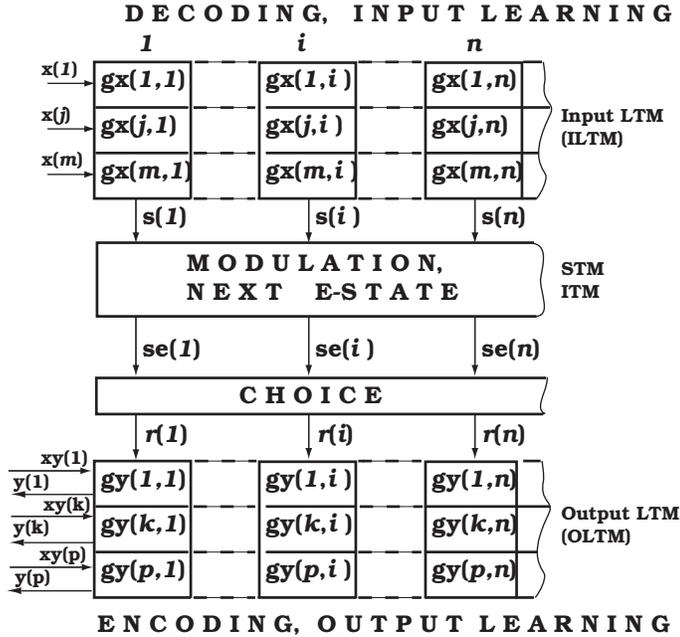}
 \end{center}
 \caption{Example of a primitive E-machine } \label{fig5}
\end{figure}

Figure \ref{fig5} illustrates the architecture of a simple PEM.  The
interpretation procedure $f_y$  is divided into four elementary
procedures: DECODING, MODULATION, CHOICE, and ENCODING.
 The model uses a single  E-state array, $e(i)$.  The
next E-state procedure, $f_e $, is similar, in this example, to the
\emph{fast-charging-slow-discharging} of a capacitor. The model
employs a universal learning algorithm, $f_g$, that simply
tape-records input/output associations in the LTM. The recorded
association gets an elevated level of residual excitation to produce
an effect of \emph{recency}. This is described as the addition to
the next E-state procedure.
 \\\\
\textbf{Variables}:
\begin{itemize}
\item $\nu \in \{0,1,...\}$ is a discrete time (the cycle number).\footnote{
 Treated as real-time cognitive models, E-machines can be thought to have a
  \emph{psychological time step} $\Delta t$
 on the order of $1-10msec$.  More complex models of E-machines with multi-step cycles
 may
 use several time variables, $\nu1$, $\nu2$,... with different time steps.}

\item $x(:)\doteq x(1),..x(m)$ is the input vector with $m$
components.  In this example, each component is treated as a symbol.
That is, only the equal/not equal relationship is defined. As
before, we use a special empty symbol, $\varepsilon$, to indicate
``no data''.

\item $wx(:)\doteq wx(1),..wx(m)$, where $wx(j) > 0$ $(j=1,..m )$ is
the vector describing the weights of input symbols.

\item $gx(:,i)\doteq gx(1,i),..gx(m,i)$ is the vector stored in the
$i$-th location of the Input LTM (ILTM), where $i\in \{1,...n\}$.

\item $s(:) \doteq (s(1),..s(n))$ is a \emph{similarity array}. In
general, $s(i)$ is a nonnegative real number representing a
similarity between $x(:)$ and $gx(:,i)$. In this example, we use a
very simple criterion of similarity -- the number of matching
non-empty symbols. Accordingly $s(i) \in \{0,1,...m\}$.

\item $e(:) \doteq (e(1),..e(n))$ is an \emph{E-state array}.
Variable  $e(i)$ is a nonnegative real number that represents the
level of \emph{residual excitation} associated with the $i$-th
location of the LTM. In this example, we use a single  E-state
array. In more complex models, several E-state arrays, $e1(:)$,
$e2(:)$,..., with different dynamic properties can be used.

\item $se(:)\doteq (se(1),..se(n))$ is \emph{modulated} or
\emph{biased} \emph{similarity array}. In general, $se(i)$ is an
element of a real array that describes the similarity affected by
the residual excitation.

\item $r(:)\doteq (r(1),..r(n))$ is a \emph{retrieval array}. In
general, $r(i)$ is an element of a real array that represents the
level of activation of the $i$-th location of OLTM. In this model,
we use a random \emph{winner-take-all} choice, so only one component
of this array, $r(iwin)$, corresponding to the winner, $iwin$, is
not equal to zero. Formally, in this example we need only the
variable $iwin$. The $r$-array is introduced for the sake of
completeness. It does not appear in the following equations. This
array is needed in more complex models of primitive E-machines that
employ more complex encoding procedures.

\item $gy(:,i)=gy(1,i),..gy(p,i)$ is the vector stored in the $i$-th
location of the Output LTM (OLTM). In this model, components of
$gy(:,i)$ are treated as symbols.

\item $y(:)=y(1),...y(p)$ is the output vector retrieved from  OLTM.
In this model the output is read from the winner location of OLTM.
Components of $y(:)$ are treated as symbols.

 \item $xy(:)$ is the input to the OLTM used for writing output data
 in  this memory.

\item $wptr$ and $wen$ are the auxiliary variables used to describe
the tape-recording learning algorithm. They   serve as the
\emph{write\_pointer} and the  \emph{write\_enable}, respectively.
 \end{itemize}
\noindent \textbf{Parameters:} \begin{itemize}

\item $a<0.5$ is a parameter that determines the modulating  effect
of $e(i)$ on $s(i)$ that produces biassed similarity $se(i)$.

\item $c<1.0$ is a parameter that determines the rate of decay of
$e(i)$. The time constant of decay is $\tau=1/(1-c)$, so
$c=1-1/\tau$.

\item $m$ is the number of components in the input vector $x(:)$.
\item $n$ is the number of locations in the ILTM and OLTM. It is the
size of
 all arrays with index $i$. Namely, $s(i)$,$se(i)$,$e(i)$, $gx(:,i)$, and $gy(:,i)$.
\item $p$ is the number of components in the output vector $y(:)$.

 \end{itemize}
\emph{Remark.}  In the experiments discussed in this paper, weights,
$wx(:)$, will be treated as parameters. In more complex experiments
the pair $(x,wx)$ can be treated as an input.

\medskip
 \noindent \textbf{Procedures}:
 \medskip

 {\sl{DECODING} (computing similarity):}
\\\\
$for$ $i=1:n$  \ \ \ (this \emph{for}  is applied to expressions
(1), (2), and (5))
 \\\\
 (1) \hspace{0.5in} $s(i)=\sum_{j=1}^m{wx(j) \cdot T(x(j)=gx(j,i) \neq \varepsilon)}$
 \\\\
 where, \ $T(z)=1$, if $z=true$, else $T(z)=0$

\medskip
 {\sl{MODULATION} (computing biased similarity):}
 \\\\
(2) \hspace{0.5in} $se(i)=s(i)\cdot (1+ a \cdot e(i))$

\medskip
 {\sl{CHOICE }(randomly selecting a winner):}
  \\\\
 (3) \hspace{0.5in}  $ iwin \ :\in \ \{i: se(i)=max(se)> 0 \}\
 \doteq  \  \textbf{MSET}$
 \\\\
 where $:\in$ denotes the operation of the random equally probable choice of an element
 from a set.

 \medskip
  {\sl{ENCODING} (retrieving data from OLTM):}
  \\\\
  (4) \hspace{0.5in}  $y(:)=gy(: \ ,iwin)$

\medskip
 {\sl{NEXT E-STATE PROCEDURE} (dynamic reconfiguration):}
 \\\\
 (5) \hspace{0.5in} $if \ (s(i)>e(i))\  e(i)(\nu+1)=s(i)\ ; \  else \ e(i)(\nu+1)=c \cdot e(i);$ \ $end$
 \\\\
 where, $e(i)(\nu+1)$ is the value of $e(i)$ at the next moment $\nu+1$.
 For simplicity, we
 do not write $\nu$  in the current values of the variables. That is, $e(i)$
 is the same as $ e(i)(\nu)$,
 $s(i)$ is the same as $s(i)(\nu)$, etc.

  \medskip
  {\sl{NEXT G-STATE PROCEDURE} (learning)}
 \\\\
 (6) \hspace{0.5in} $if(wen==1)\  gx(:,wptr)(\nu+1)=x(:);\  gy(:,wptr)=xy(:)$; \\
  . \hskip1.9in$wptr(\nu+1)=wptr+1;$ $end$

\medskip
\emph{Remark.} Expression (6) tape-records input and output
vectors in the  ILTM and OLTM, respectively, when recording is
enabled, $wen=1$. For simplicity, in this model, we assume that the
weights of input symbols, $wx(:)$, do not affect recording. We will
always have $wx(j)\geq 1.0$.

\medskip

{\sl{ADDITION to the NEXT E-STATE PROCEDURE} (the recorded location of LTM, \\
 $i=wptr$, gets initial residual excitation)}
\\\\
 (7) \hspace{0.5in} $if(wen==1)\  e(wptr)(\nu+1)= s(wptr)= \sum_{j=1}^m{wx(j)
 \cdot T(x(j) \neq \varepsilon)};$ $end$

\medskip
 \emph{Remark.} The truth function $T(z)$ is defined in
expression (1).
  Expression (7) adds residual excitation to the location
 of ILTM with $i=wptr$, -- the location in which data was just recorded. This
 happens only if recording is enabled, $wen=1$. The
 level of added  residual excitation, $e(wptr)$, is  equal to the one that would be produced if the input
  vector $x(:)$ were already recorded in this location of ILTM, that is, as if
  $gx(:,wptr)=x(:)$. This trick allows one to avoid introducing
  intermediate steps in the $\nu-th$ cycle.

\medskip
\noindent \textbf{System of references.} The model described above
will be referred to as Model (\ref{sec5.2}) or PEM (\ref{sec5.2}),
meaning the model (PEM) described in Section \ref{sec5.2}.
Similarly, expression (\ref{sec5.2}.6) will mean expression (6) of
Section \ref{sec5.2}, etc.

 \subsection{On neural implementation of primitive E-machines} \label{sec5.3}

\begin{figure}[h!] \begin{center}
  \includegraphics[width=3.5in]{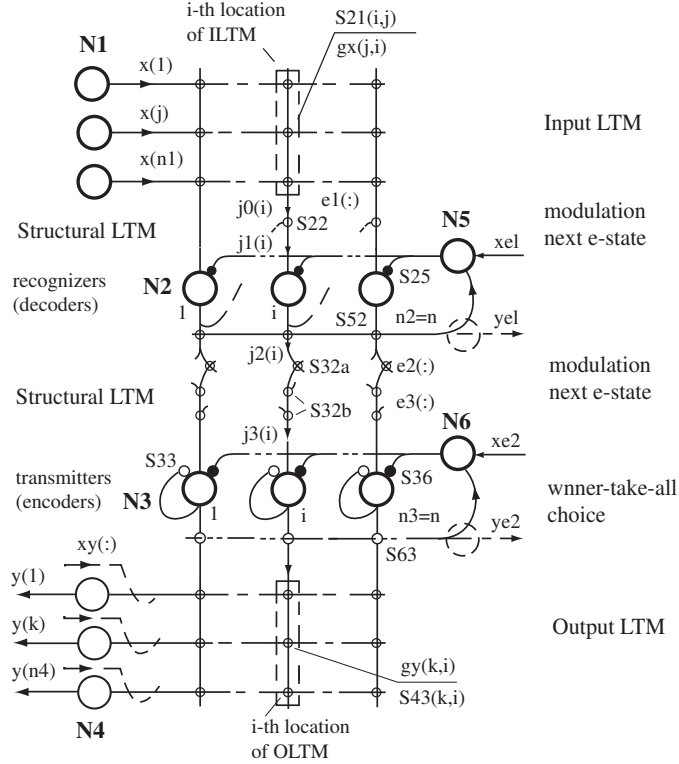}
 \end{center}
 \caption{An example of a neural implementation of a primitive E-machine} \label{fig6}
\end{figure}

It is helpful to have an intuitive link between E-machines and
neural networks. Such a link provides a source of neurobiological
heuristic considerations for the design of the models of E-machines
and the source of psychological heuristic considerations for the
design of the corresponding class of neural models. Figure
\ref{fig6} shows the general architecture of a homogeneous neural
network corresponding to a PEM. The architecture was discussed in
\cite{bib8, bib10}.

 Large circles with incoming and outgoing lines represent \emph{centers} --
 elements that are assigned certain coordinates in the network. A
center with incoming and outgoing lines can be interpreted as a
neuron  with its dendrites and axons, respectively, or a network
functionally equivalent to a  ``large neuron''. Small circles
represent \emph{couplings} -- the elements whose position in the
network is described by a pair of  centers communicating through
this coupling. A coupling can be interpreted as a synapse or a
circuit functionally equivalent to a ``large synapse''. The white
and the black small circles represent excitatory and inhibitory
synapses, respectively. In what follows we
 use the terms neuron and synapse instead of the
terms center and coupling, respectively \footnote{ We use the term
\emph{coupling} rather than the traditional term \emph{connection}
to emphasize that, in general,  we treat a synapse as a complex
nonlinear dynamical element with E-states, rather than just a
connection with a variable weight.}.

Figure \ref{fig6} displays six sets of neurons (N1, N2, N3, N4, N5,
and N6) and 10 sets of synapses (S21, S22, S25, S52, S32a, S32b,
S33, S36, S63, S43).

Nj(i) is the $i$-th neuron from the $j$-th set. Skja(n,m) is the
synapse between neurons Nj(m) and Nk(n), where 'a' is an additional
index describing the type of synapse (in case there are several
different types of synapses between the sets of neurons Nj and Nk).
The ':' substituted for an index indicates the whole subset of
elements (variables) corresponding to the entire set of values of
this index. S21(i,:)$\doteq$(S21(i,1),...S21(i,n1)). S21(:,:) is the
same as S21, etc.

The diagram depicts three types of synapses with E-states:
$e1(:)$,$e2(:)$, and $e3(:)$, serving different purposes. $e2(:)$ is
similar to the $e(:)$ of Model \ref{sec5.2}; e1(:) describes the
effect of lateral modulation that allows the network to decode
temporal sequences; e3(:) describes spreading residual excitation
producing the effect of  broad temporal context. \textbf{Note}. This
neural network corresponds to a PEM more complex than Model
(\ref{sec5.2}) \cite{bib8, bib10}.

The Input LTM and the Output LTM are implemented, respectively, as
the input, and output synaptic matrices, $S21(:,:)$ and $S4(:,:)$.
The network also has some intermediate synaptic memory called
Structural LTM (SLTM). This memory corresponds to modifiable
connections among neuron decoders, N2, and neuron encoders, N3.

Layer N3 implements a \emph{winner-take-all} (WTA) choice. The layer
has local excitatory feedbacks, S33, and a global inhibitory
feedback via neuron N6 -- the, so-called,
\emph{inhibit-everyone-and-excite-itself} principle. The WTA choice
could also be implemented by using inhibitory synapses S33(i2,i1)
with  gains $\geq 1.0$ for all synapses except those with $i2=i1$ --
the, so-called, \emph{inhibit-everyone-but-itself} principle.

Layer N2 has an inhibitory feedback with the gain less than unity to
provide contrasting. Both layers, N2 and N3, have inhibitory inputs
xe1, and xe2. The diagram also depicts some outputs ye1, ye2
carrying information about the global levels of activation of the
corresponding layers. One can invent different functional models of
primitive E-machines corresponding to the network of Figure
\ref{fig6}. However, an attempt to discuss this interesting topic in
more detail would take us too far from the goal of this paper.

\medskip \noindent \emph{Remarks}:

1. To implement very large ILTM and OLTM one needs intermediate
neurons to increase the fan-out of input neurons N1, and the fan-in
of output neurons N4. Some ideas inspired by the organization of the
cerebellum were discussed in \cite{bib8}.

\medskip 2. The concept of E-machine supports the notion that the
E-states (the states of dynamic STM and ITM)  are associated with
the properties of individual neurons and synapses. There is an
interesting possibility to formally connect the dynamics of the
phenomenological E-states with the statistical conformational
dynamics of ensembles of membrane proteins treated as Markov systems
\cite{bib12}.

\medskip 3. Some neural network researchers strongly oppose the
notion of \emph{symbols in the brain}. We argue that this
anti-symbolic bias is counterproductive.  There is an interesting
possibility to represent symbols in ILTM and OLTM as sparse
quasi-orthogonal synaptic vectors (see items 7, 8, and 9 in Section
\ref{sec9}).

\medskip 4.  It is challenging to try to find neurobiological
mechanisms that could provide an implementation of learning
algorithms functionally close to the complete memory algorithm used
in Model (\ref{sec5.2}). Such mechanisms may exist, taking into
account the principle limitations of the traditional learning
algorithms demonstrated in Section \ref{sec4}.

\section{A PEM can learn to simulate a GRAM} \label{sec6}

\textbf{Theorem 3.} The PEM described in Section \ref{sec5.2} used
as system AS in the experimental setup of Figure \ref{fig4} can
learn to simulate a GRAM from a sample of the GRAM's behavior of the
length $\geq n \cdot m$.
 \\\\
\emph{Proof}. First of all, we need to specify the parameters of the
PEM (\ref{sec5.2}) and the conditions of the experiments.

System AS is organized as Model (\ref{sec5.2}) with parameters
$AS.m=2$; $AS.p=1$; \\ $AS.wx(1)=AS.wx(2)=1.0$. We assume that
$AS.n$ is as big as needed to record all training data.

\medskip

\emph{Remark}. We add the name AS to the names of parameters of
Model (\ref{sec5.2}), to avoid confusion with the parameters of the
GRAM. The names $m$ and $n$ refer to the parameters of GRAM.

 \medskip

The experiment is divided into two parts: \emph{training}, and
\emph{examination}. Training starts at $\nu=0$ and lasts until
$\nu=\nu1-1$. During training $NS.sel=1$, that is, $AS.xy=dout$.
Writing to LTM is enabled, $AS.wen(\nu)=1$. At the beginning of
training, the write pointer is in the initial position,
$AS.wptr(0)=1$, and the LTM is empty, $AS.gx=AS.gy=\varepsilon$, and
residual excitation is set to zero, $AS.e(:)=0$. We assume that
during training each fixed rule of the GRAM is recorded at least
once in the LTM. Since the number of such rules is $n \cdot m$, we
need $\nu1\geq n\cdot m$.

 Examination starts at $\nu=\nu1$ and lasts as long as needed. During examination
$NS.sel=0$, that is, $AS.xy=AS.y$. For simplicity we assume that
writing to LTM is disabled, $wen(\nu)=0$. \textbf{Note}. Writing
could be enabled during examination. This would make no difference
as far as Theorem 3 is concerned.
\\\\
 We begin with a verbal explanation of the \emph{effect of
working memory }. The effect is produced by decaying residual
excitation, $e(i)$, associated with locations of LTM.

Let the $i$-th location of the LTM (ILTM and OLTM) contain the
record of the following fixed rule: $gx(1,i)=a, gx(2,i)=d,
gy(1,i)=d$, where $a \in \textbf{A}$ and $d \in \textbf{D}- \{
\varepsilon \}$. The residual excitation $e(i)$ associated with this
location can reach the maximum possible level, $emax=2$, in two
situations:

\begin{enumerate}
 \item At the next moment after the rule was
recorded in the $i$-th location. This is the result of
 expression (7) from Section (\ref{sec5.2}) (expression (\ref{sec5.2}.7))
 which sets residual excitation at the moment of recording.
 \item If the rule is already recorded and input is $x(1)(\nu)=a$, and $x(2)(\nu)=d$.
 In this case, $e(i)(\nu+1)=2$ is a result of
 expressions (\ref{sec5.2}.1) and (\ref{sec5.2}.5).
\end{enumerate}

Once the $e(i)=emax$ is set, we can say that the rule $(a,d,d)$ is
\emph{placed in working memory}. The reason for this statement is
that, if we send input $x(1)=a$ $x(2)=\varepsilon$ the output $y(1)$
will be retrieved from one of the locations of LTM with the highest
level of residual excitation among locations for which $gx(1,i)=a$.
Locations for which $gx(1,i) \neq a$ will have $s(i)=0$ due to
expression (\ref{sec5.2}.1). Accordingly, due to expression
(\ref{sec5.2}.2), for these locations $se(i)=0$ independently of the
level of residual excitation. This effect of executing the most
recent rule placed in working memory will last until $e(i)$ decays
below a certain level, $eloss$. For this model $eloss=1$ and  the
modulating coefficient, $a$, in expression (\ref{sec5.2}.2) must be
less than $1/emax=.5$.  Due to expression (\ref{sec5.2}.5), the time
of decay of $e(i)$ from $emax$ to $eloss$ is
\\\\
(1)  \hspace{.10in} $tmax\ = \ \ln(eloss/emax)/\ln(c)
 = -\ln(emax/eloss) \cdot \ln(1-1/\tau) > \ \tau \cdot ln(emax/eloss)$
\\\\
where $c =1-1/\tau$, and, for this model, $emax/eloss=2$.  We can
transform this qualitative explanation into a rigorous proof by
verifying the following statements:

\begin{description}
\item \textbf{S1}. At $\nu=\nu1$ (the beginning
of examination) LTM of AS contains at least once any fixed rule
             $(addr,din,dout) \in \textbf{A} \times \textbf{D}' \times
             \textbf{D}'$, where  $\textbf{D}' = \textbf{D}-\{\varepsilon\}$.  Formally,
\\\\
 (2)  \hspace{.10in} $\forall (addr,din,dout) \in  \textbf{A} \times \textbf{D}' \times
             \textbf{D}'
  \  \exists i (gx(1,i),gx(2,i),gy(1,i))= (addr,din,dout))$
\\\\
This statement follows from expression (\ref{sec5.2}.6) and the
definition of the experiment of training. During training $wen=1$,
so the training sequence containing all fixed rules is tape-recorded
in LTM.
 \item \textbf{S2}. At $\nu=\nu1$ all location of LTM
containing fixed rules have $e(i)>0$ due to expression
(\ref{sec5.2}.7). Let $tmax > \nu2-\nu0$. From expression (1) we
find that  to guarantee that $e(i)= emax=2$ will not decay
 below $eloss$ during the entire experiment
 of training and examination it is sufficient to have $tmax\geq
 \nu2-\nu0$.  This condition is satisfied if
 \\\\
  (3)  \hspace{.10in} $\tau\geq\ \tau min
  \doteq(\nu2-\nu0)/ln(eloss/emax)\ =\ (\nu2-\nu0)/ln2$

 \item \textbf{S3}. Suppose in the course of examination we test AS only in the
 read mode. That is, we send only inputs
 $addr,din$ with $din=\varepsilon$. Then $y(1)$ will be retrieved from a
 location $ird$ for which $gx(1,ird)=addr$ and $e(ird)$ has the
 highest level among all locations with $gx(1,i)=addr$. If $\tau\geq \tau min$, it is
 guarantied that $e(ird)>eloss$, so $y(1)$ will
 be retrieved from  the location in which a certain data,
 $ din \neq \varepsilon$ was recorded most recently and, therefore, $y(1)=din$.
 This is exactly how GRAM would react to this input.

 \item \textbf{S4}. Let $\nu w$ and $\nu r$, where $\nu r >\nu w$ be some
 ``write'' and ``read'' moments in the course of
 examination, such that $x(1)(\nu w)=addr$, $x(2)(\nu w)=din \neq \varepsilon$
 and $x(1)(\nu r)=addr$, $x(2)(\nu r)= \varepsilon$.
 Let there be no other input between $\nu w$ and $\nu r$ with the same $x(1)$
 and different $x(2)\neq \varepsilon$. Due to
 expression (\ref{sec5.2}.5) $e(i)(\nu w+1)=emax=2$
 for all $i \in \{ i : gx(1,i)=addr, gx(2,i)=din \}$. Accordingly, due to what was said
 in \textbf{S3}, $y(1)(\nu r)=din$. Again, AS reacts exactly as GRAM.
 \end{description}
 This proves the theorem, since all other possible cases
 of ``writing to working memory'' and reading from this memory can be
 reduced to \textbf{S3} and \textbf{S4}.

\medskip

\noindent \emph{Remark}. Why, in this model should the data loss
level, $eloss$,  be set at 1? The reason for that is that the
residual excitation $e(i)=1$ can be created in a wrong location by
$x(2)=din \neq \varepsilon$. Accordingly, only the residual
excitation higher than 1 can be relied upon. The $e(i)>1$  can only
be the result of decay from level emax=2. The latter level can be
set only by the input $x(1)=addr$, $x(2)=din \neq \varepsilon$. This
input corresponds to writing non-empty data, $din \neq \varepsilon$,
to an address, $addr$, of GRAM.

\section{On Turing universality and learning} \label{sec7}

We have already shown that system AS of Figure \ref{fig1}, organized
as PEM (\ref{sec5.2}), can learn to simulate a GRAM with $n$
locations and $m$ data symbols in $n \cdot m$ steps. To make the
brain, $B=(NS,NM,AS,AM)$, a Turing-universal learning system, we
need a system AM capable of learning to simulate the finite-state
part of a Turing machine (\ref{sec5.2}).

Because of the delayed speech feedback, NM3.\emph{y} $\rightarrow$
AM.\emph{x2}, it is sufficient for AM to be a learning system
universal with respect to the class of combinatorial machines.  The
PEM (\ref{sec5.2}) satisfies this requirement.

\textbf{Theorem 4.} System AM, organized as PEM (\ref{sec5.2}), can
be taught, in the experimental setup of Figure \ref{fig1}, to
simulate, in principle, any combinatorial machine with 2-inputs and
3-outputs. The above  statement holds for any $m\geq 1$ and $p\geq
1$.
 \\\\
\emph{Proof}. Let $AM.m=2$, $AM.p=3$, $AM.wx(:)=1.0$, $AM.a=0$. The
latter means that modulation is turned off, and, according to
Expression (\ref{sec5.2}.2),  $AM.se(i)=AM.s(i)$.
 \\
 As in Section
\ref{sec6}, we divide the experiment into two stages: training and
examination.

At the stage of training $NM.sel=NS.sel=1$. The input vector, is
coming from system (W,D), that is, $AM.x(1)=dout$, and the output
vector is coming from the teacher, that is,
$AM.xy(:)=NM.y(:)=T.y(:)$. Training starts at $\nu=0$ and ends at
$\nu=\nu1-1$. At the beginning of training the LTM is empty,
$AM.(gx,gy)(0)=0$, and the write pointer is in the initial position,
$AM.wptr(0)=1$. Writing is enabled during training, $AM.wen(\nu)=1$
for $\nu=0,...\nu1-1$.

At the stage of examination, $\nu \geq \nu1$, writing is disabled,
$AM.wen(\nu)=0$, and $NM.sel(\nu)=0$, that is the output, $NM.y(:)$
is produced by system AM, $NM.y(:)=AM.y(:)$. The teacher controls
select signal $NS.sel$ and can switch it between the external
system, (W,D), and the imaginary external system, the role of which
is played by system AS.

Since, at the stage of training, the teacher controls all motor
signals $NM.y(:)$, any set of productions $AM.(gx(1,i),gx(2,i))
\rightarrow AM.(gy(1,i),gy(2,i),gy(3,i))$ can be formed in the LTM
of system AM. It is easy to verify that these productions will be
correctly interpreted (executed) by Expressions
(5.2.1),(5.2.2),(5.2.3), and (5.2.4). Expression (5.2.2) is reduced
to $se(i)=s(i)$ since $a=0$, and, consequently, Expressions (5.2.5)
and (5.2.7) have no effect. This completes the proof.

\medskip

\noindent \emph{Remarks:}

1. To transform the system of Figure \ref{fig1} into a working model
one needs to take care of the synchronization of units
(W,D),NS,NM,AS, and AM. These technical problems are of no
significance for the purpose of this paper. The model was actually
implemented as an interactive C++ program for the MS Windows called
EROBOT.

\medskip 2. The model has an interesting psychological
interpretation. It suggests that the Turing universality of human
thinking is a result of mental interaction of the motor system, AM,
that forms $Sensory,Motor\rightarrow Motor$ ($SM\rightarrow M$)
associations, and the sensory system, AS, that forms $MS\rightarrow
S$ associations. The symbols M and S in the left parts of the
associations suggest motor and sensory feedbacks. Using the
terminology borrowed from \cite{bib4}, system AM can be called the
\emph{central executive}. It has a speech feedback that produces the
effect of the internal states (the states of mind) of the simulated
Turing machine. The robot needs to \emph{talk to itself} to be able
to perform, in principle, arbitrarily complex computations.

 \section{The effect of a context-dependent mental set}
\label{sec8}

So far we have demonstrated the importance of the E-states in the
sensory system AS. The introduction of the E-states in the motor
system AM  leads to an interesting effect of context-dependent
mental set. We are going to show that the PEM (\ref{sec5.2}) used as
AM can acquire  the knowledge needed to simulate $N=2^{2^m}$
$m$-input Boolean functions during only $n=2^{m+1}$ steps of
training. The effect is achieved due to a context-dependent dynamic
reconfiguration of this limited knowledge at the time of decision
making. Let $m=10$. Then $n=2048$ and $N=2^{1024}$. In the shown
case, k=2, n=8, and N=16.

\begin{figure}[h!] \begin{center}
  \includegraphics[width=4.75in]{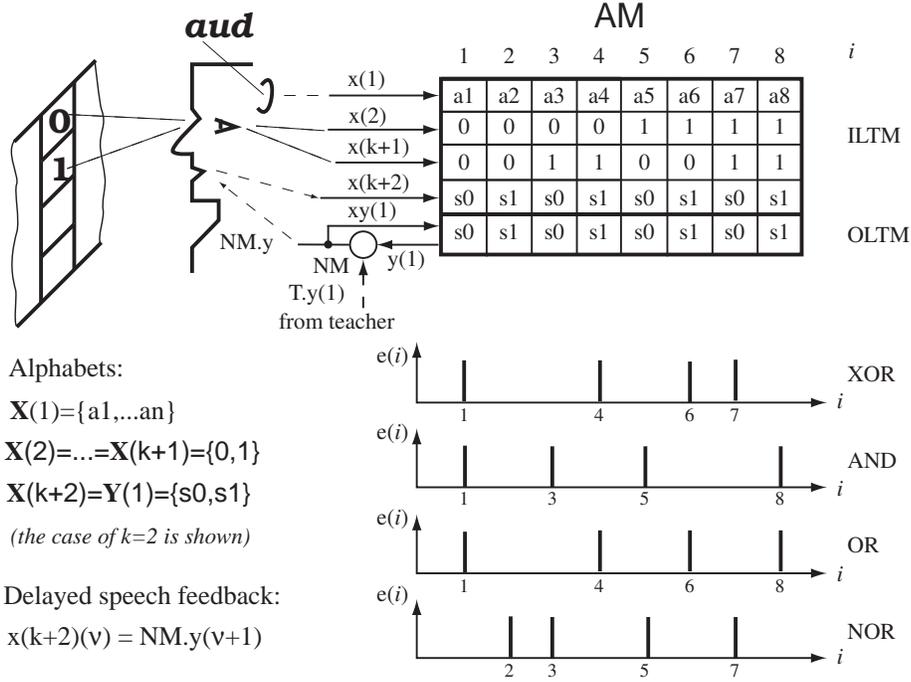}
 \end{center}
 \caption{Experimental setup for demonstrating the effect of context-dependent mental set.
 } \label{fig7}
\end{figure}
 Let us modify the experimental system of Figure
 \ref{fig1} as shown in Figure \ref{fig7}.  We  connect
 the auditory channel, $aud$, and allow the robot to see several
 squares at a time. In the discussed experiment, the robot does not need to type on the
 keyboard. It  expresses its reactions via the speech channel. We
 removed system AS and centers NS. These units are not needed in the
 discussed experiment.

 Let AM be described as PEM (5.2) with $m=k+2$, and $p=1$. In Figure
 \ref{fig7}, $k=2$. The table displays the contents
 of LTM after $n=2^{k+1}=2^3=8$ steps of training.  The upper part
 of the  table shows the contents of ILTM, $gx(1:k+2,1:n)=gx(1:4,1:8)$. The
 lower part shows the contents of OLTM, $gy(1,1:n)=gy(1:8)$. Each
 association, $gx(:,i)\rightarrow gy(1,i)$, has  $k+2=4$
 input symbols and one output symbol. To
 understand the structure of an association  consider the association at
 $i=3$. The upper symbol $gx(1,1)=a3$ represents an auditory name
 assigned to this association. The name is taken from the set
 (alphabet) of auditory names  $\textbf{X}(1)=\{a1,...a8\}$.

 Symbols $gx(2:3,3)=(0,1)$ represent the visual input read from the screen. The symbols are taken from
 the visual alphabets $\textbf{X}(2)=..\textbf{X}(3)=\{0,1\}$
In this experiment,  the robot can see $k=2$ visual symbols
simultaneously through inputs $x(2:3)$. At the moment shown, the
robot sees symbols $x(2:3)=(0,1)$.

The output symbol, $gy(1,3)=s0$ represents the motor speech symbol
corresponding to the reaction $0$ -- the robot simulates a Boolean
function with the input read from the screen, and the output
represented as a motor speech symbol. The motor speech symbols  are
taken from the alphabet $\textbf{Y}(1)=\{s0,s1\}$. The diagram
depicts a delayed proprioceptive speech feedback,
$x(k+2)(\nu)=NM.y(:)(\nu+1)$. At $i=3$, the corresponding
proprioceptive symbol is stored as $gx(k+2,3)=s0$. The role of the
proprioceptive speech feedback is explained later.

In what follows we give a verbal explanation of the work of the
model. The explanation can be transformed into a rigorous proof in
the same way it was done in Section \ref{sec6}. As before, the
experiment is divided into two stages: \emph{training} and
\emph{examination}. We assume that the model has a RESET button that
allows the experimenter to set $e(:)=0$, and a GO/STOP button to
start and stop execution. We also assume that the speech feedback
can be turned ON and OFF. If it is OFF, $x(k+2)=\varepsilon$. We
start with the case when the feedback is OFF, so for a while we
disregard $gx(k+2,i)$ and $x(k+2)$.

At the beginning of  training, LTM is empty, $gx,gy=\varepsilon$, and
the write pointer is in the initial position, $wptr=1$. During
training, writing is enabled, $wen=1$, and the teacher creates
associations corresponding to all productions participating in all
$k$-input Boolean functions. The number of such productions, and the
minimum length of training, is $n=2\cdot 2^k$. In the shown case,
$n=2\cdot2^2=8$. Each production is given a unique auditory name
from $\{a1,..an\}$.

At the stage of examination, writing is disabled, $wen=0$. Using
auditory input x(1), the experimenter pre-tunes (pre-activates) a
set of associations corresponding to a given Boolean function by
sending the auditory names of these associations via input $x(1)$.
Let all $wx(j)=1.0$ for $j=2,..k+1$, and $wx(1)>k+1$. In this case
the maximum level of residual excitation, $e(i)$, created via the
auditory channel will be bigger than the maximum level of excitation
created via all other channels. Since the time of decay of residual
excitation can be made arbitrarily big, any number of productions
can be put into working memory (cf  Section \ref{sec6}). While
residual excitation of the selected subset of associations stays
above the loss level, $eloss$, the robot will execute the pre-tuned
Boolean function. For the discussed model $eloss=k$.

We can now explain the idea of the speech feedback. Suppose each
input cycle is divided into two steps. At the first step, the robot
executes the association as if there is no feedback. At the second
step, proprioceptive input $x(k+2)$ arrives and increases the level
of residual excitation of the executed association. Then the
feedback is interrupted, and the next cycle starts. And so on.  This
\emph{refresh} mechanism will produce the effect of a
\emph{self-supporting mental set}. (One needs to slightly modify the
learning algorithm and the training procedure to correctly record
the proprioceptive symbols into $gx(k+2,wptr)$.

  \section{Promising directions of research} \label{sec9}

There are many interesting problems and possibilities associated
with the development, exploration, and understanding of the
symbolic-dynamical computational paradigm represented by E-machines
\cite{bib8}. These include:

\medskip 1. \textbf{Decoding temporal sequences}. Adding lateral
 pre-tuning to the next E-state
 procedure addresses this problem. The corresponding PEM can learn
 to simulate, in principle, any output independent finite memory
 machine. Introducing a delayed feedback in the above PEM leads to a system capable of
 learning to simulate any output dependent finite memory machine.

\medskip 2. \textbf{Waiting associations}. One can add an E-state
creating  an effect of waiting association.
 In this way one can  produce an effect of a finite stack without
 using a RAM buffer, and simulate context-free grammars with limited memory.
 This allows one to get an effect of calling and returning from
 subroutines.

\medskip 3. \textbf{Broad temporal context}. This effect can be
produced by adding  different types of ``spreading'' E-states.
Changing the time constants and the radii of spread of such states
leads to the effects of mental set significantly more complex than
that discussed in Section~\ref{sec8}.

\medskip 4. \textbf{Active scanning of associative memory}. People
can answer the questions about what happened after or before  a
certain event. This effect can be achieved by first creating an
E-state profile that activates the information about the mentioned
event, and then actively shifting this  profile in the time-wise or
counter-time-wise direction. This can be done by introducing
additional control inputs in a PEM.

\medskip 5. \textbf{Inhibiting data with given features}. One does
not have to think about the E-states as ``excitations'' or
``activations''. One can imagine the E-states producing inhibiting,
pre-inhibiting, pre-activating, post-activating, etc. effects. In
fact, any functions over sets of data stored in the LTM that assign
some dynamic labels to the subsets of data can be thought of as
different kinds of E-states. For example, one can imagine a
situation in which one first creates an E-state activating a set of
data (by addressing this data by content) and then temporarily
inhibits the selected set of data by sending some inhibiting control
input. The brain has many different neurotransmitters and receptors
that may justify different hypotheses about the possible types of
such control inputs. From the viewpoint of this paper, the important
thing is that different hypotheses of this type can be efficiently
expressed in terms of the E-machine formalism.

\medskip 6. \textbf{Connecting the dynamics of E-states with the
statistical conformational dynamics of ensembles of membrane
proteins}. There is an interesting possibility of connecting some
E-states with the statistical conformational dynamics of the
ensembles of membrane proteins treated as \emph{probabilistic
molecular machines} (PMM) \cite{bib11,bib12}. A PMM is a Markov
system with the transitional probabilities depending on macroscopic
inputs such as membrane potentials, and concentrations of
neurotransmitters \cite{bib15, bib25}. With this approach, an
ensemble of PMMs (EPMM) becomes a statistical mixed signal computer
in which the E-states can be interpreted as the occupation numbers
of different microscopic states of PMMs. The formalism is a natural
system theoretical extension of the classical Hodgkin and Huxley
(HH) theory \cite{bib16}, and works well for the generation of
spikes \cite{bib12}. Validating this approach for representing the
dynamics of short-term and intermediate-term memories  would justify
postulating quite complex next E-state procedures.

\medskip
\noindent\emph{Remark}. There can be  many different
hypotheses as to how the E-states of different types could be
implemented at the cellular level \cite{bib18,bib19}. Translating
such neurobiological hypotheses into next E-state procedures gives
one a mathematical tool for exploring psychological implications of
these hypotheses.

\medskip 7. \textbf{Effects of drugs on higher mental functions}. As
shown in Section 8, changing dynamical E-states changes the symbolic
personality of an E-machine. Accordingly, the E-machine formalism
gives one a tool for expressing a broad range of ideas about how
different drugs could affect higher mental functions by affecting
the E-states.

\medskip 8. \textbf{From signals to symbols and vice versa}. The
most likely candidate for the E-machine paradigm is the neocortex.
To make this paradigm practically implementable at the higher level,
the lower levels of the brain must be able to convert sensory
signals into sensory symbols, and motor symbols into motor signals.
One of the possibilities is that the \emph{signal-to-symbol}
conversion is done by different feature detectors. Each detector
would have a fixed sparse ID serving as a unique pointer to this
detector. The sets of such IDs would form primary sensory alphabets.
Similarly, the units generating primary motor features would have
sparse IDs forming primary motor alphabets. With this approach, the
higher association areas of the neocortex could deal mostly with
symbols.

\medskip 9. \textbf{Sparse recoding in a hierarchical structure of
associative memory}. As mentioned before, a complex E-machine (CEM)
is a hierarchical associative learning system built from several
PEMs. The concept of sparse IDs can be extended to allow the PEMs of
higher levels to store data in terms of sparse-recoded references to
the data stored in the PEMs of lower levels. This would naturally
produce different effects of data compression and chunking
\cite{bib1}.

\medskip 10. \textbf{Communication among PEMs via association
fibers}. The sparse encoding of symbols addresses the question of
how  the PEMs of different modalities and levels could efficiently
communicate via association fibers. The numbers of association
fibers are not big enough to provide crossbar connectivity.  In the
case of sparse encoding of symbols, several symbolic messages could
be sent simultaneously with a low level of crosstalk. This would
allow any small subset of talkers to broadcast their sparse-encoded
messages simultaneously to large numbers of listeners. The only
limitation is that not too many talkers must talk at the same time
through the same set of association fibers. As an example, our
estimate shows that around $m\approx 50$ neuron-talkers can talk
simultaneously to $n\approx 10^9$ neuron-listeners through $k\approx
10^4$ association fibers.  That is, \emph{any-n1-to-any-n2}
communication is physically impossible if $n1= n2 = 10^9$. However,
\emph{any-m-of-n1-to-any-n2} communication is quite possible if
$m\approx 50$, and $n1= n2 \approx 10^9$.

\medskip 11. \textbf{Computing statistics on the fly depending on
context}. At the symbolic level, the brain may not pre-compute
statistics at the time of learning because statistics depends on
context.  \emph{How could the neocortex compute statistics on the
fly depending on context?} A sparse encoding of symbols offers a
solution to this problem. If we change the procedures of CHOICE and
ENCODING to allow several sparse-encoded symbols to be read
simultaneously from the locations of OLTM with a ``high enough''
level of activation, summing up several sparse vectors produces a
statistical filtering effect \cite{bib8}.

\medskip 12. \textbf{The problem of a natural language}. This is the
most challenging and interesting problem that, we believe, is well
suited for the integrated symbolic-dynamical approach discussed in
this paper. (See \cite{bib17} to learn about the problem.)  The
effect of context-dependent mental set, illustrated in Section
\ref{sec8}, reveals the tip of an iceberg. It takes less information
to dynamically activate the data structures that are already present
in the LTM than to create new data structures. Even less information
is needed if some data structures are already pre-tuned through the
inputs of other modalities. Accordingly, unlike the statements of a
formal language, the sentences of a natural language do not need to
carry complete information. A hint can be sufficient to remove
ambiguity in a given context. This sheds light on why people with
similar backgrounds (G-states) and mental sets (E-states) can
efficiently communicate via short messages, whereas people with
different backgrounds and mental sets have difficulties
understanding each other.

\emph{How can different effects of language generation and
understanding be produced without a conventional RAM buffer?}

\medskip 13. \textbf{On emotions and motivation}. We should mention
the problem of emotions and motivation. \emph{Can the E-machine
paradigm shed light on this problem?} Let us postulate that, at a
higher level, there  exists some symbolic representation of emotions
-- otherwise our language would not have the names for our emotional
states. If this postulate is correct, the E-machine formalism can be
applied to the higher level learning involving emotions. People
remember their pleasant and unpleasant emotional states. This means
that, at a higher level, the effect of positive/negative
reinforcement cannot be reduced to the effect of
increasing/decreasing the weights of sensorimotor associations. We
postulate that, at a higher level, the brain forms associations
involving the symbols representing emotions and other observable
\emph{internal} (I) states. The sets of these associations can be
dynamically reconfigured depending on context by changing the
E-states. This helps to understand why our concepts of ``good'' and
``bad'' depend on our knowledge and  mental set.

It is easy to imagine a situation when retrieving emotional symbols
affects control inputs that change the E-states that, in turn,
affect the retrieval of emotional symbols, and so on. This would
shed light on the nature of various self-reinforcing loops, such as
the well known panic attack loop.

 \section{Methodological remarks} \label{sec10}

\subsection{How complex is the untrained human brain?}
\label{sec10.1}

 Let (W,D,B) be a cognitive system, where (D,B) is a human-like robot
with the basic cognitive characteristics similar to those of a
person. Let B(t) be a formal representation of the robot's brain at
time t, where t=0 corresponds to the beginning of learning (an
\emph{untrained} brain). Let, for the sake of concreteness,
$t=t_{20}\doteq 20 \ years$ represent a \emph{highly trained}
(intelligent) brain. The specific number is not important. It could
be  10, or, perhaps, even 5 years. We argue that B(0) must have a
relatively short formal representation (megabytes), whereas
B($t_{20}$) must have a very long representation (terabytes). That
is, methodologically, it is advantageous to look first for the
representation of B(0), and then try to understand how B(0) changes
into B(t) (with bigger and bigger t) in the course of learning.

\medskip

\noindent \emph{Remark}. One of the reasons  a representation of
B(0) cannot be too long is that this representation must be encoded,
in some form, in the human genome. The whole genome takes about
$700MB$ -- not enough room for B($t_{20}$). One may argue that the
size of B(0) must include not only the size of the genetic code, but
also the size of the procedure that translates the genetic code into
B(0). It is reasonable to postulate that the size of the latter
procedure is significantly smaller than the size of B(0). After all,
the same cellular machinery creates biological systems of vastly
different complexity from different genetic codes.

\subsection{Cognitive science as a physical theory} \label{sec10.2}

Theoretically, a consistent mathematical cognitive theory must be
able to derive arbitrarily complex cognitive phenomena from a model
of (D,B(t)) interacting with the external world, W. The situation
can be loosely compared with that in a traditional physical theory.
To get a specific metaphor, consider the problem of simulating the
behavior of the electromagnetic field in a Linear Accelerator (LAC).
The mathematical model underlying the latter simulation can be
represented as a pair (M,C), where M are the Maxwell equation and C
are the specific constraints (boundary conditions and sources)
describing the design of LAC. In this case, it is quite obvious that
it would be impossible to simulate the specific behavior of the
system (M,C) without having an adequate representation of the basic
constraints M. We argue that it is similarly impossible to simulate
specific cognitive phenomena in system ((W,D),B(t)) without having
an adequate representation of the ``basic constraints'' B(t).

\medskip

\noindent\emph{ Remark.} It is not unusual for the whole physical
phenomenon to have a simpler mathematical representation than its
parts. For example, the whole behavior of an electromagnetic field
has an efficient formal representation (the Maxwell equations).
However, in the case  of nontrivial external constraints, it is
practically impossible to find separate formal representations for
either the electric or the magnetic projections of this behavior.
The concept of E-machine suggests that the same holds for the
symbolic-dynamical behavior of the human brain. The whole behavior
has a simpler representation than its symbolic and/or dynamical
projections.

\subsection{The problem of an adequate formalism} \label{sec10.3}

What was said in Sections \ref{sec10.1} and \ref{sec10.2} raises the
problem of an adequate mathematical formalism for representing (and
thinking about) different levels of the physical phenomenon of
information processing in the human brain. There is a big difference
between a \emph{numerically correct} mathematical representation and
an \emph{adequate} mathematical representation. An adequate
representation must allow our brain to efficiently think (mentally
simulate) the phenomenon in question. To understand the problem
consider two extreme examples.

At one extreme, imagine a model of a PC with the MS Windows
operating system represented as a system of billions of nonlinear
differential equations with some very complex initial conditions.
Note that, in this ``dynamical'' model, the software would be
represented as initial conditions. With some encoding of symbols as
real vectors, and with sampling some vectors at the right moments,
the model would produce correct simulation results. It is safe to
say that nobody (except the designer of the model) would be able to
understand how this dynamical model works.

At another extreme, imagine a model of the behavior of the
electromagnetic field in LAC represented as a Turing machine
simulating the discussed behavior with, say, 256 bit accuracy. Given
enough time and enough memory (tape) space, this ``symbolic'' model
would produce numerically correct simulation results. As before,
nobody (except the designer) would be able to understand how this
symbolic model works.

The fact is that our brain cannot efficiently think about
\emph{symbolic} computations in \emph{dynamical} terms and about
\emph{dynamical} computations in \emph{symbolic} terms. Therefore,
an adequate mathematical formalism for representing and thinking
about the physical phenomenon of information processing in the human
brain must take this fact into account. It does not help to know
that any physically implementable computing system, including the
brain, can be represented, in principle, in either symbolic or
dynamical terms. We need to figure out what system we need to
represent, rather than how to represent a given system.

Finding an adequate formal representation of B(0) is not the main
challenge.  A much greater challenge is to develop an adequate
language (a system of metaphors and mental models) that would allow
us (humans) to understand the behavior of system (W,D,B(t)) with
larger and larger values of t. Even in the case of simple
 Maxwell equations, the behavior of the pair (M,C) from
Section \ref{sec10.2} can be extremely complex. The complexity of
the behavior comes from the complexity of the specific external
constraints, C. The same must be true in the case of an adequate
mathematical theory of system (W,D,B(t)). Trying to put too many
specific constraints in B(0) inevitably reduces the predictive power
of the cognitive theory.

\subsection{On system integration and falsification} \label{sec10.4}

\emph{If it is true that B(0) has a relatively short formal
representation, why do we have difficulties reverse engineering this
representation?}

We argue that many basic principles of organization and functioning
of B(0) are already known. What is missing is an adequate
formalization, extrapolation, and integration of these basic
principles into a single  mathematical theory of the whole human
brain as an integrated computing system. The critical issues for the
development of such a system level theory are \emph{system
integration} and \emph{falsification}. The following considerations
explain why these issues must not be separated from each other.

Let $c_1,..c_m$ be some basic properties of the human brain as an
integrated computing system, and let $\textbf{C}_i$ be the set of
all possible systems with the property $c_i$. If one treats
$c_1,...c_m$ as constraints on a single integrated model of B(0),
 the search area for B(0) is the intersection  of sets
$\textbf{C}_1,...\textbf{C}_m$. The more properties one considers,
the smaller becomes the search area. Systems outside this area are
eliminated from the search by the falsification principle. In
contrast, if one ignores the issues of system integration and
falsification and treats each property, $c_i$, independently (just
as a biological inspiration for the development and study of systems
from $\textbf{C}_i$), the search area is the union of
$\textbf{C}_1,...\textbf{C}_m$. In this case, the more properties
one considers, the bigger becomes the search area.  Consequently, it
becomes increasingly difficult to ``eliminate the impossible'' and
find the truth. (The terminology is borrowed from the famous Sir
Arthur Conan Doyle quotation: \emph{"When you eliminate the
impossible, whatever remains, however improbable, must be the
truth."})

As the users of similar brains, we have an unlimited source of
reliable system-level constraints on the whole human brain as an
integrated computing system. We do not have (and may never be able
to obtain) similarly reliable constraints on
 the parts of the brain and/or the parts of the brain's performance.
 Therefore, methodologically, it is important not to separate the problem of
 the parts of the brain and the  parts of the brain's behavior from the problem
 of the whole brain.

 Recently, there has been a resurgence of interest in the
 whole brain (especially the neocortex) as an integrated computing
 system \cite{bib14a, bib14b, bib31, bib20}.

\subsection{On dumb learning and smart interpretation}
\label{sec10.5}

In Section 4, we have shown that losing information at the time of
learning leads to principal limitations at the time of decision
making. Theoretically, no system can learn more than a system with a
complete memory algorithm (CMA). As the E-machine formalism
demonstrates, a powerful enough interpretation procedure can  make
up for a dumb but universal learning algorithm. In contrast, no
interpretation procedure can make up for a smart learning algorithm
that loses information. The popular notion of a smart learning
algorithm contradicts to the requirement of universality of human
learning, and creates a methodological pitfall -- no fixed learning
algorithm can be smart enough to know in advance what information
may turn out to be important in the future.

In terms of the $AM \leftrightarrow AS$ architecture of Figure
\ref{fig1}, all traditional learning systems can be characterized as
models of AM. It is difficult to define explicitly what AM must do.
Accordingly, it is difficult to rigorously demonstrate the
limitations of the traditional learning systems as brain models. The
situation becomes more transparent when one treats learning systems
as models of system AS rather than AM. The properties of many
interesting external systems, (W,D), can be formally defined.

In Section \ref{sec4}, we used this approach to show that many
traditional learning systems cannot learn to simulate external
systems, (W,D), with the properties of a read/write memory, whereas
the human brain can. This learning problem -- call it the
\emph{RAM-buffer-problem} -- presents a harder falsification test to
the neural (connectionist) theories of learning than did the famous
\emph{XOR-problem} \cite{bib18a}.

 \medskip
\noindent \emph{Remark.} It should be emphasized that postulating
the existence of a conventional RAM buffer would not explain  how
the brain learns to simulate external systems with the properties of
a read/write memory. It would remain absolutely unclear how it could
learn to use such a conventional RAM buffer to simulate different
memory aids. That is, the RAM buffer behavior must be learned. In
psychological terms, this means that a brain, B, not exposed to the
external system, (W,D), with the properties of a read/write memory
would not learn to use its working memory, and, consequently, would
not develop an IQ needed to perform nontrivial mental computations.
 \\\\
\textbf{Disclaimer}. The system level constraints discussed in this
paper (such as Theorems 1 and 2 of Section \ref{sec4}) are
applicable to the higher (symbolic) level of human learning. They
introduce no limitations on the learning algorithms that can be used
at the lower levels (as long as these algorithms do not lose
important information needed at the higher levels). In item 8 of
Section \ref{sec9}, we postulated that the lower levels perform
\emph{signal-to-symbol} and \emph{symbol-to-signal} transformations.
We have not attempted to discuss the  important question of what
type of transformations can be performed at these levels. Evolution
has found a large number of efficient signal processing solutions.
Which parts of these solutions are genetically determined and which
are affected by learning is an open question.

\subsection{Are there principle limitations on what can be learned?}
\label{sec10.6}

Some theories of learning equate the problem of learning with the
problem of deciphering the structure of a target machine (teacher)
observed as a black box by another machine (learner). Usually the
target machine is treated as a grammar that has to be identified
from the set of sentences \cite{bib14}. With this definition of
learning, the learner cannot learn to simulate the behavior of the
teacher of the type higher than type 3 (finite-state grammars) --
even not all type 3 behaviors are learnable. This general result
seems to contradict to Theorem 3 of Section \ref{sec6} showing that
the behavior of a GRAM can be learned -- a GRAM is a system of type
0.

In fact, there is no contradiction. System AS in Figure \ref{fig4}
(learner) does not treat GRAM (teacher) as a black box. Defined as
PEM (5.2), AS has a built-in information about the external system
(W,D).  Accordingly, the black box limitations  do not apply. We
argue that the same holds for the phenomenon of human learning. A
human learner does not treat a human teacher, and other external
systems  as black boxes. It expects these systems to have certain
properties.

As  shown in Section \ref{sec7}, with such a ``grey box'' approach
to learning, there are no principle limitations on what can be
learned. It is an experimental question (a falsifiable hypothesis),
as to whether this approach is applicable to the problem of human
learning.

\medskip \emph{Remark}. What was said above suggests that the
falsification principle should be applied to the formulations of the
biologically-inspired mathematical problems, not just to the
solutions of these problems. As is well known, to adequately define
a physical (natural) problem in mathematical terms is not easier
than (is essentially the same as) to solve this problem. This
includes biological problems as well. The catch is that biological
problems are already defined by Nature. Accordingly, a great
challenge for a biologically-consistent mathematical theory is to
decipher and adequately formalize these natural definitions -- not
to replace them with  our own artificial definitions.



\begin{thebibliography}{34}

\bibitem{bib1} Anderson, J.R. (1976). Language, Memory, and Thought.
Hillsdale, New Jersey: \emph{Lawrence Erlbaum Associates,
Publishers.}


\bibitem{bib3} Baddeley, A.D. (1982). Your memory: A user's guide.
\emph {MacMillan Publishing Co., Inc.}

\bibitem{bib4} Baddeley, A.D. (1992). Working Memory. SCIENCE, VOL.
255. 556:559.


\bibitem{bib5} Chomsky, N. (1956). Three models for the description
of language. \emph{I.R.E. Transactions on Information Theory}. JT-2,
113-124.

\bibitem{bib6} Collins, A.M., and Quillian, M.R., (1972). How to
make a language user. In E. Tulving and W. Donaldson (Eds.)
Organization and memory. \emph{New York: Academic Press}.

 \bibitem{bib7} Eliashberg, V. 1967. On a class of learning machines. \emph{Moscow:
Proceedings of VNIIB, \#54}, 350-398.

\bibitem{bib8} Eliashberg, V. (1979). The concept of E-machine and
the problem of context-dependent behavior. \emph{TXU 40-320, US
Copyright Office}.

\bibitem{bib9} Eliashberg, V. (1981). The concept of E-machine: On
brain hardware and the algorithms of thinking. \emph{Proceedings
 of the Third Annual Meeting of Cognitive Science Soc.}, 289-291.

\bibitem{bib10} Eliashberg, V. (1989). Context-sensitive associative
memory: ``Residual excitation'' in neural networks as the mechanism
of STM and mental set. \emph{Proceedings of IJCNN-89, June 18-22,
1989, Washington, D.C.} vol. I, 67-75.

\bibitem{bib11} Eliashberg, V. (1990). Molecular dynamics of
short-term memory. \emph{Mathematical and Computer modeling in
Science and Technology.} vol. 14, 295-299.

 \bibitem{bib12} Eliashberg, V.
(2005). Ensembles of membrane proteins as statistical mixed-signal
computers. \emph{Proceedings, IJCNN 2005.}


\bibitem{bib14} Gold, E.M. (1967). Language identification in the
limit. \emph{Information and Control, 10:447-474.}

\bibitem{bib14b} Grossberg, S. (2006). Towards a unified theory of
the neocortex. Technical Report CAS/CNS TR-2006-008. .

\bibitem{bib14a} Hawkins, J. and Blakeslee, S. (2004). On
Intelligence. \emph{Times Books.}

\bibitem{bib15} Hille, B. (2001). Ion Channels of Excitable
Membranes. \emph{Sinauer Associates, Inc}.

\bibitem{bib16} Hodgkin, A.L. and Huxley, A.F. (1952) A quantitative
description of membrane current and its application to conduction
and excitation in nerve.  Journal of Physiology, 117(4): 500-544.



\bibitem{bib17} Feldman, J.A. (2006). From Molecule to Metaphor. A
Neural Theory of Language. \emph{A Bradford Book, the MIT Press.}

\bibitem{bub17a} Johnson, R.C. (1989). `E-states' mediate cognition.
Electronic Engineering Times, January 2, 1989, pp. 67,68,90.

\bibitem{bib18} Kandel, E.,R. (2006). In Search of Memory. The
Emergence of a New Science of Mind. \emph{W.W. Norton and
 Co.}
\bibitem{bib18a} Minsky, M. \& Papert, S. (1969). Perceptrons.
\emph{Cambridge MA: MIT Press}.

\bibitem{bib18b} Minsky, M. (1988). The Society of Mind.  Simon \&
Schuster Inc.

 \bibitem{bib19} G. Mongillo, O. Barak, M. Tsodyks. (2008).
Synaptic Theory of Working Memory. SCIENCE, VOL. 319, 1543:1546.

 \bibitem{bib20} M. de Kamps, V. Baier, J. Drever, M.Dietz, L. M\"{o}senlechner, F. Vander Velde.
 (2008). The state of MIND. Neural Networks. 21, 1164-1181.







\bibitem{bib25} Nichols, J.G., Martin, A.R., Wallace B.G., (1992)
 From Neuron to Brain, \emph{Third Edition, Sinauer Associates}.

\bibitem{bib26} Rumelhart, D.E., McClelland, J.L. (Eds.) (1986).
Parallel Distributed processing: Explorations in the Microstructure
of Cognition. \emph{Cambridge, MA: MIT Press.} (Vols 1 and 2.)

\bibitem{bib27} Sun, R., Alexandre, F. (Eds.) (1997).
Connectionist-symbolic integration. From Unified to Hybrid
Approaches. \emph{Lawrence Erlbaum Associates.}


\bibitem{bib29} Smolensky, P., Legendre, G. 2006. The Harmonic Mind:
 From Neural Computation To Optimality-Theoretic Grammar Vol. 1:
Cognitive Architecture; vol. 2: Linguistic and Philosophical
Implications. \emph{MIT Press.}

 \bibitem{bib30} Turing, A.M. (1936). On computable
numbers, with an application to the Entscheidungsproblem.
\emph{Proc. London Math. Society}, ser. 2, 42.



\bibitem{bib31} Widrow, B., Aragon, J.C.  (2005). "Cognitive"
memory.  Proceedings. IEEE International Joint Conference on Neural
Networks, Volume 5, 3296 - 3299.


 \bibitem{bib33} Zopf, G.W. (1961). Attitude and
Context. In ``Principles of Self--organization''. \emph{Pergamon
Press}, 325-346.
 \bibitem{bib34} Eliashberg, V., Eliashberg, Y. (2007). The Mathematics of the Brain.
 Proposal for
 DARPA/DSO SOL, DARPA Mathematical Challenges, BAA 07-68. Mathematical Challenge One.
 Grant, Award No. FA9550-08-1-0129.


\end{thebibliography}
\end{document}